\documentclass[letterpaper, 10 pt, conference]{ieeeconf}  % Comment this line out if you need a4paper

\IEEEoverridecommandlockouts                              % This command is only needed if 
\usepackage{amssymb}  % assumes amsmath package installed

\usepackage{glossaries}
\usepackage{booktabs}
\usepackage{tablefootnote}
\usepackage{multirow}
\usepackage{bm}
\usepackage{xurl,hyperref}

\usepackage{todonotes}
\setuptodonotes{inline}

\makeglossaries
\newacronym{hsi}{HSI}{Hyperspectral Imaging}
\newacronym{hdc}{HDC}{Hyperdimensional Computing}
\newacronym{hcv}{HCV2}{Hyperspectral City V2}
\newacronym{oa}{OA}{overall accuracy}
\newacronym{aa}{AA}{average accuracy}
\newacronym{miou}{mIoU}{mean intersection over union}
\newacronym{ppv}{PPV}{\textit{proportion of positive values}}

\newcommand{\ie}{\textit{i.\,e.}}
\newcommand{\eg}{\textit{e.\,g.}}
\newcommand{\etal}{\textit{et al.}}
\newcommand{\justoliunet}{1D-Justo-LiuNet}
\newcommand{\hyko}{HyKo2}
\newcommand{\hsidrive}{HSI-Drive}
\newcommand{\fone}{\textnormal{F}_1}
\newcommand{\hypso}{HYPSO-1}
\newcommand{\hdcmr}{HDC-MiniROCKET}
\newcommand{\mr}{MiniROCKET}
\newcommand{\negspace}{-2.5em}

\title{\LARGE \bf
Data-Efficient Spectral Classification of Hyperspectral Data Using \mr\ and \hdcmr\
}

\author{Nick Theisen$^{*}$, Kenny Schlegel$^{\dagger}$, Dietrich Paulus$^{*}$ and Peer Neubert$^{*}$% <-this % stops a space
\thanks{This work was partially funded by Wehrtechnische Dienststelle 41 (WTD), Koblenz, Germany}% <-this % stops a space
\thanks{The authors denoted with $^{*}$ are with the Institute of Computational Visualistics, University of Koblenz, Germany. Correspondence Email: {\tt\small nicktheisen@uni-koblenz.de}}%
\thanks{The authors denoted with $^{\dagger}$ are with the Chair of Automation Technology, TU Chemnitz, Germany}%
\thanks{© 2025 IEEE.  Personal use of this material is permitted.  Permission from IEEE must be obtained for all other uses, in any current or future media, including reprinting/republishing this material for advertising or promotional purposes, creating new collective works, for resale or redistribution to servers or lists, or reuse of any copyrighted component of this work in other works.}%
}

\begin{document}

\maketitle
\thispagestyle{empty}
\pagestyle{empty}

%%%%%%%%%%%%%%%%%%%%%%%%%%%%%%%%%%%%%%%%%%%%%%%%%%%%%%%%%%%%%%%%%%%%%%%%%%%%%%%%

\begin{abstract}
The classification of pixel spectra of hyperspectral images, \ie\ spectral classification, is used in many fields ranging from agricultural, over medical to remote sensing applications and is currently also expanding to areas such as autonomous driving. Even though for full hyperspectral images the best-performing methods exploit spatial-spectral information, performing classification solely on spectral information has its own advantages, \eg\ smaller model size and thus less data required for training. Moreover, spectral information is complementary to spatial information and improvements on either part can be used to improve spatial-spectral approaches in the future. Recently, 1D-Justo-LiuNet was proposed as a particularly efficient model with very few parameters, which currently defines the state of the art in spectral classification. However, we show that with limited training data the model performance deteriorates. Therefore, we investigate MiniROCKET and HDC-MiniROCKET for spectral classification to mitigate that problem. The model extracts well-engineered features without trainable parameters in the feature extraction part and is therefore less vulnerable to limited training data. We show that even though MiniROCKET has more parameters it outperforms 1D-Justo-LiuNet in limited data scenarios and is mostly on par with it in the general case. 
\end{abstract}
\section{Introduction} \label{sec:intro}

% Motivation

\Gls{hsi} can capture the spectrum of light reflected by surfaces in many very narrow bands -- up to hundreds of spectral channels. These high-dimensional spectra are a valuable basis to derive certain material properties or discriminate surface materials better than possible with RGB-images.
The concepts of multi- and  \gls{hsi} in remote sensing dates back to the 1970s \cite{Landgrebe2002HID}. Hyperspectral images are often recorded from satellites, planes or UAV plattforms, but processed on the ground because of hardware constraints. However, in recent years multiple works were published that examine small and efficient models that can be deployed on the platform itself to prioritise data transmission from satellites to ground stations through limited bandwidth communication channels, \eg\ \cite{Pitonak2022CFB,Justo2024SSI,Kovac2024DLF}.

Recently, a particularly efficient model, named \justoliunet\ was proposed in \cite{Justo2024SSI}. It was designed to be deployed on a satellite which observes the earth with a hyperspectral camera. The model assesses what image regions are covered in clouds and therefore do not need to be transferred to earth. The authors compared it against other efficient models and showed that it achieves state-of-the-art performance for spectral classification on multiple remote sensing datasets with only a few thousand parameters. 

\begin{figure}
    \centering
    \includegraphics[width=\columnwidth]{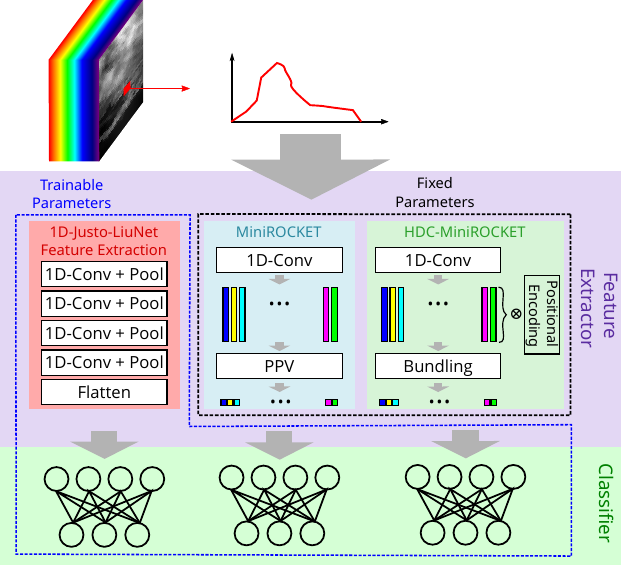}
    \caption{Overview of the models for spectral classification that we compare in this work. Each pixel of a hyperspectral image provides a reflection spectrum. \justoliunet\ uses training data to fit the parameters of the feature extractor. \mr\ and \hdcmr\ do not need labeled training data to fit their feature extractor, making them more data-efficient and in general more stable in situations with only very limited training data. (Details in Sec. \ref{sec:models}).}
    \label{fig:overview}
\end{figure}

Following this direction, one can think of ground applications where only narrow communication channels are available or where a certain degree of autonomy on a platform for interaction with its surroundings is required, \eg\ probing areas after disaster events when only ad-hoc communication infrastructure is available or precision agriculture applications where plants health status could be assesed in-situ and sent to a central data platform. Furthermore, data-efficient spectral classification models can be used to complement spatial information.

A typical task in \gls{hsi} is semantic segmentation, \ie\ given a hyperspectral image predict which class out of a given set of classes each pixel belongs to. Related to this is spectral classification, \ie\ the classification of a measured reflection spectrum of some surface. Spectral classification can be understood as a special case of semantic segmentation where each 1-dimensional pixel spectrum is processed individually.

Today, the best results for semantic segmentation are achieved by models that process spatial and spectral information simultaneously using 2D- and 3D-CNNs or transformer architectures. However, spectral classification has certain advantages: First, models have in general less parameters and can therefore often be trained with less data. Second, samples for semantic segmentation models are represented by images or image patches, but most \gls{hsi} datasets currently consist of only a low number of images. Samples for spectral classification models are spectra, \ie\ each pixel of a hyperspectral image with corresponding label is a training sample. Of such, a much higher number of training samples are available. Third, the features do not mix spectral with spatial information, which simplifies investigation of which patterns in the data have the most influence on classification results. Finally, there are applications in which it is sufficient to capture a single sample with a reflection spectrometer instead of a hyperspectral camera, \eg\ estimating the chlorophyll content in leaves \cite{Flores2024RCK}. Spectrometers also have the advantage that they are cheaper, easier to calibrate and they can be combined with a calibrated light source to provide well-defined illumination \cite{Geladi2007MIH}.

Collection and annotation of spectral samples often takes expert knowledge and can be expensive. Hence, data-efficient models are required for spectral classification.
\justoliunet\ has only few parameters, but its feature extraction still requires labeled data for training. In contrast, \mr\ \cite{Dempster2021MAV} uses a fixed set of engineered convolutions to extract features. Hence, it is a promising candidate for spectral classification in a limited data setting.

Schlegel \etal\ \cite{Schlegel2022HET} showed that \mr\ is a special case of \hdcmr, which uses \gls{hdc}-mechanisms to encode positional relations of elements in the input sequence in the final feature vector, which would otherwise be lost (see \ref{sec:models:rocket}). The position of an element in a spectral sequence carries semantics, because it represents a specific spectral band. Therefore, \hdcmr\ is a promising extension to \mr\ for spectral classification.

In summary, our contributions are as follows:
\begin{itemize}
    \item We investigate \mr\ for spectral classification and compare it to \justoliunet\ -- the current state-of-the-art model -- and show that \mr\ achieves comparable results in the general case on three datasets. 
    \item We further show that \mr\ is well-suited for spectral classification in limited data setting. It outperforms \justoliunet\ when the train data is reduced below a certain threshold and suffers less from bias toward majority classes when datasets are imbalanced.
    \item Finally, we experiment with \hdcmr\ to better capture the positional relationships between bands in a spectral sequence. To our knowledge, \hdcmr\ was not used for spectral classification, yet. However, we did not observe a clear improvement with the \gls{hdc}-extension in our experiments.
\end{itemize}

\section{Related Work}
In \cite{Theisen2024HAB} it was shown that the models that use spectral-spatial features for \gls{hsi} semantic segmentation benefit from reducing the spectral dimension and using pre-trained data. They conclude that to train such models insufficient data is available. Hence, efficient spectral classification models -- such as \justoliunet\ \cite{Justo2024SSI} --  if they can be trained with few  training samples, could complement models that use spatial information and mitigate the problem of limited \gls{hsi} training data. 

\mr\ \cite{Dempster2021MAV} was proposed as an efficient approach for time-series classification (see \ref{sec:models:rocket}) and is a promising candidate for spectral classification as stated in sec. \ref{sec:intro}. It is generally applicable for classification of 1-dimensional signals. In the following we provide an overview over usage of \mr\ in the context of hyperspectral data.
Flores \etal\ \cite{Flores2024RCK} estimate chlorophyll content in plant leaves from reflectance spectra in the spectral range of 400 - 800nm wavelength. They compare the regression performance of \mr, a 4-layer neural network, XGBRegressor and RandomForest and achieve best results with a regressor built on \mr\ features.
In \cite{Silva2024ETG} the authors compare different models, \ie\ CNN, ResNet, InceptionTime and \mr\ for sugar content estimation in wine grape berries. They show that the tested models are able to generalize to unseen grape varieties and vintages. While \mr\ did not result in the lowest RMSE values among all models, it showed satisfactory performance with only a fraction of the parameters and significantly lower computation time than all other models \cite{Silva2024ETG}.
Another application  for \mr\ is proposed in \cite{Yan2025AOM} where it is used to estimate pollutant gas concentations in hyperspectral samples collected from satellites in the infrared spectrum fast and accurately.

\section{Algorithmic Approaches to Spectral Classification} \label{sec:models}

This section introduces the models that were used in our experiments. The input to those are spectral vectors, \ie\ a 1-dimensional signal $\pmb{x} \in \mathbb{R}^{b}$ consisting of a sensor response for each of the sensors $b$ spectral bands. The output is an activation vector $\pmb{y} \in \mathbb{R}^{c}$ corresponding to the probability that a sequence belongs to each of $c$ classes. Table \ref{tab:datasets} gives an overview of the data characteristics. Typically, all models use a single fully connected layer with softmax activations as a classifier. The main difference lies in the feature extraction modules.

\subsection{\justoliunet}

The LiuNet architecture consists of multiple layers of 1d-convolutions followed by max pooling and was proposed in \cite{Liu2018TLF}. The feature extractor consists of four layers with width 32, 32, 64 and 64. The resulting features are flattened and used to train a classifier. \cite{Justo2023AOH} and \cite{Justo2024SSI} use LiuNet as basis to define \justoliunet. The final model was designed to trade-off performance and efficiency in terms of low amount of parameters. \justoliunet s convolutional layers have a width of 6, 12, 18 and 24 and use a kernel size of 6, resulting in only around 4,500 trainable parameters for the \hypso\ land-sea-cloud dataset.

\subsection{ROCKET models} \label{sec:models:rocket}

The ROCKET family of algorithms was developed for time series classification and evolved over time. Their feature extractors do not rely on labeled training data \cite{Dempster2020REF,Dempster2021MAV,Schlegel2022HET}.

First in line, was the ROCKET algorithm, which was proposed by Dempster \etal\ \cite{Dempster2020REF}. It has only a single convolutional layer, which consists of 10 000 randomly generated convolutional kernels of different size, dilation factor, weights, biases and padding. These are used to convolve the input signal to highlight corresponding patterns. The resulting 10 000 signals are compressed with global max pooling and \gls{ppv} and the result is flattened to a 20 000 dimensional feature vector. The \gls{ppv} results describe how apparent a certain pattern corresponding to a convolutional kernel is in the input signal. See \cite{Dempster2020REF} for details. 

\mr\ is a more efficient and mostly deterministic version of the ROCKET algorithm proposed in \cite{Dempster2021MAV}. Instead of using random convolutions, \mr\ uses a predetermined and mostly deterministic set of engineered kernels for feature extraction. Further, it exploits some proporties of the algorithm to significantly improve runtimes. In this context using max global pooling for feature compression does not provide relevant information and is therefore omitted. The length of the sequences defines the dilation sizes, \ie\ shorter sequences require less dilation, and for each combination of kernel and dilation the bias is set based on the output of a convolution of a randomly selected input sequence with the corresponding kernel. \mr\ transforms the input sequences into 9996-dimensional feature vectors and was shown to be up to $75\times$ faster than ROCKET on large datasets \cite{Dempster2021MAV}. See \cite{Dempster2021MAV} for details.

In \cite{Schlegel2022HET} Schlegel \etal\ showed that a dataset can simply be constructed in which \mr\ is not much better than random guessing. They show that the problem can be resolved by maintaining positional relationships of the elements in the input sequence during feature extraction. They achieve this by converting the intermediate features (after the convolution) to bipolar, which allows it to use \gls{hdc} mechanisms to \textit{bind} ($\bigotimes$) a positional encoding vector to the feature vectors of each element in the input sequence. Furthermore, conveniently the results of the \gls{ppv} operation and the \gls{hdc} \textit{bundling} operation ($\bigoplus$) are proportional. See \cite{Schlegel2022HET} for details. With this solution positional relationships can be kept during feature extraction elegantly with only minimal computational overhead.
\section{Experimental Setup}

In this section we first introduce the datasets used for evaluation and then provide details on the setup and implementations. Experimental results will be presented and discussed in sec. \ref{sec:results}.

\subsection{Datasets} \label{sec:datasets}
In our evaluation, we use multiple datasets. First, \hypso\ sea-land-cloud \cite{Justo2023AOH} that Justo \etal\ use in \cite{Justo2024SSI} to show the superiority of \justoliunet\ against a wide range of other models. In their evaluation the authors additionally use two datasets to estimate the models generalization capability, \ie\ data captured during the EO-1 mission with cloud annotations \cite{Kovac2024DLF} and annotated satellite imagery of Dubai\footnote{ \url{https://humansintheloop.org/resources/datasets/semantic-segmentation-dataset-2/} (Last visited: 23.06.2025)}. We omit the former because labels are not publicly available and the latter because it consists of RGB images and our focus lies on \gls{hsi}. 

Second, to compare the models for a broad range of applications, we use \hyko\ \cite{Winkens2017Hyko} and \hsidrive\ \cite{Basterretxea2021HSIDrive} -- from the HS3-Bench benchmark \cite{Theisen2024HAB}, which consists of driving scenes. HS3-Bench \cite{Theisen2024HAB} contains 3 datasets with spectral dimensionality ranging from 15 to 128 bands and number of classes ranging from 9 to 18 classes. However, we omit \gls{hcv} \cite{Li2022HCV} for now because its size and resolution leads to extreme training times of multiple days per run. In our experiments we use the same data splits as in \cite{Theisen2024HAB}.

The \hypso\ \cite{Justo2023AOH} dataset consists of 38 hyperspectral images collected from the HYPSO-1 satellite. Each image has a label per pixel marking it as \textit{sea}, \textit{land} or \textit{cloud}. Information on the technical details of the sensor can be found in \cite{Bakken2023HCF}. Following \cite{Justo2023AOH} we remove $8$ bands during preprocessing and normalize the data to a range of $[0,1]$. Further we employ the same train-validation-test split (30/3/5 images) in our experiments as the authors of \cite{Justo2024SSI}. Unfortunately, at the time of experimentation, the image with ID 36 was not available on the dataset website. Therefore, we used only 29 images for training.

An overview over the used datasets properties can be found in Table \ref{tab:datasets}. While the classes in \hypso\ sea-land-cloud dataset are mostly balanced, the class distributions in the HS3-bench datasets are highly imbalanced. For more detailed information see \cite{Theisen2024HAB} and \cite{Justo2023AOH}.

\begin{table}
    \footnotesize

    \caption{Overview over the datasets used for experimentation.}
    \centering
    \resizebox{\columnwidth}{!}{%
    \begin{tabular}{c|ccc}
        Name & \hyko & \hypso & \hsidrive \\\hline
        Image size & $254 \times 510$ & $956 \times 684$ & $409 \times 216$ \\
        Bands ($b$)& $15$ & $120$ & $25$ \\
        Range [nm] & $470$-$630$ & $400$-$800$ & $600$-$975$ \\
        Spectral Resolution [nm] & $15$ & $5$  & n/a \\
        Images & $371$ & $38$ & $752$ \\
        Classes ($c$) & $10$ & $3$ & $9$ \\
        Train/Test/Val-split ($\%$) & $50$/$20$/$30$ & $79$/$8$/$13$ & $60$/$20$/$20$\\
    \end{tabular}
    }
        \label{tab:datasets}
\end{table}

\subsection{Setup \& Implementation}

We use only train data for model training. Reported performance scores are gathered on the test set, if not otherwise stated. During training we export the model weights of the epoch in which the validation \gls{miou} score is maximal and use this model for evaluation. Additionally, we used the validation set to experiment with and validate hyperparameters. We repeat all experiments 5 times with different seeds and report the mean results, if not otherwise stated, to get more reliable performance estimates.

Justo \etal\ \cite{Justo2024SSI} use a batch size of $32$ and train for only $2$ epochs. This configuration slows down training as parallelization of GPU is not exploited. Instead we use a larger batch size of $4096$ pixels and compensate for the lower amount of optimization steps by training for $10$ epochs. We validate these training parameters and our implementation of \justoliunet\ by training it on \hypso\ and make sure we reach similar validation accuracy scores (around $90\%$) as reported in \cite{Justo2024SSI}. These training parameters and fixed random seeds are used for all of our experiments, if not otherwise stated. Further, we use AdamW \cite{Loshchilov2019Decoupled} without weight decay and a learning rate of 1e-3 for \justoliunet\ and 3e-5 for \mr\ and \hdcmr. The learning rates where chosen as a trade-off between stability and performance. Furthermore, we had to remove the last convolutional layer of \justoliunet\ during training on \hyko\ as it has only 15 channels which leads to problems when it is pooled 4 times.

The \hdcmr\ and \mr\ implementation is based on the \texttt{pytorch}-based version of \mr\ in \textit{tsai} \cite{tsai} and the \hdcmr\ code provided by \cite{Schlegel2022HET} and we reimplemented \justoliunet\ in pytorch. Our code is organized with \texttt{pytorch-lightning}\footnote{\url{https://lightning.ai/docs/pytorch/stable/} (Last visited: 23.06.2025)} and we use \texttt{torchmetrics}\footnote{\url{https://lightning.ai/docs/torchmetrics/stable/} (Last visited: 23.06.2025)} for performance metric calculation.

\section{Experimental Results} \label{sec:results}
\subsection{Impact of Limited Data on Model Performance} \label{sec:jlnmr}
\begin{figure}
    \centering
    \includegraphics[width=\columnwidth]{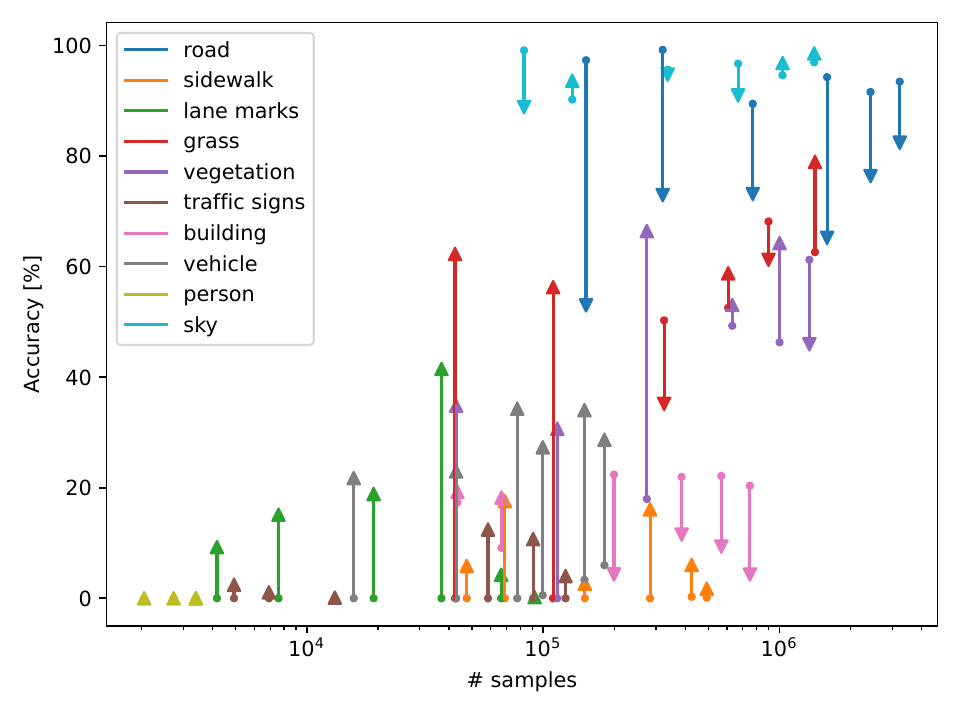}
    \vspace{\negspace}
    \caption{Difference in class accuracy between \justoliunet\ and \mr\ per number of absolute samples in train set shares for \hyko. Arrows point from \justoliunet s accuracy score towards the score for \mr.}
    \label{fig:classresults_trainratio_hyko-jln-mr}
\end{figure}

\begin{figure}
    \centering
    \includegraphics[width=\columnwidth]{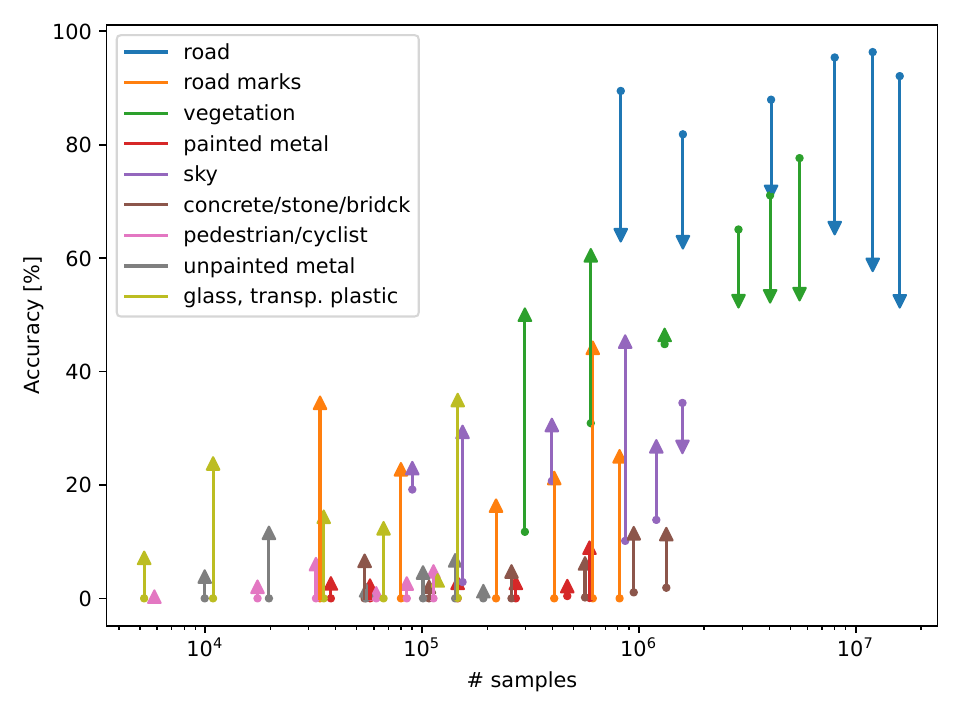}
    \vspace{\negspace}
    \caption{Difference in class accuracy between \justoliunet\ and \mr\ per number of absolute samples in train set shares for \hsidrive. Arrows point from \justoliunet s accuracy score towards the score for \mr.}
    \label{fig:classresults_trainratio_hsidrive-jln-mr}
\end{figure}

\begin{figure}
    \centering
    \includegraphics[width=\columnwidth]{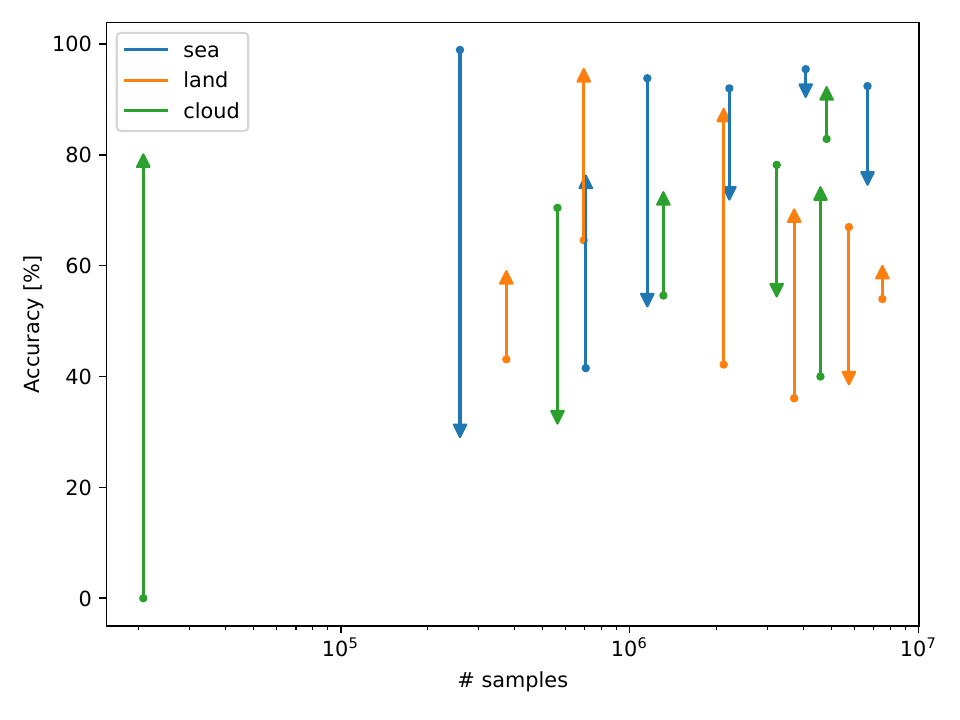}
    \vspace{\negspace}
    \caption{Difference in class accuracy between \justoliunet\ and \mr\ per number of absolute samples in train set shares for \hypso. Arrows point from \justoliunet s accuracy score towards the score for \mr.}
    \label{fig:classresults_trainratio_hypso-jln-mr}
\end{figure}

We hypothesize that with restricted access to training data \mr\ outperforms \justoliunet, as \mr s\ feature extractor does not require training. We validate this hypothesis by training both models on shares of all datasets introduced in \ref{sec:datasets}. Results are shown in Fig. \ref{fig:classresults_trainratio_hyko-jln-mr} to \ref{fig:classresults_trainratio_hypso-jln-mr}. 
 To simulate the availability of less data we train each model on a share of $p = 5, 10, 25, 50, 75, 100$ percent of all available training datasets and assess its performance on the test set. We sample full images out of the original train sets instead of single spectra to maintain a realistic class distribution, especially for the imbalanced HS3-Bench datasets. 
 
 The plots in Fig.~\ref{fig:classresults_trainratio_hyko-jln-mr} -  \ref{fig:classresults_trainratio_hypso-jln-mr} show the performance given the absolute number of samples for each class. The classes are represented in different colors. The arrows indicate the difference in accuracy score from \justoliunet\ to \mr, \ie\ upward arrows mean \mr\ is better, downward arrows mean that \justoliunet\ is better. Per class there are six arrows, each corresponding to the number of samples per training data share. 
 The results for \hyko\ in Fig. \ref{fig:classresults_trainratio_hyko-mr-hdcmr} for \hsidrive\ in Fig. \ref{fig:classresults_trainratio_hsidrive-jln-mr} and \hypso\ in Fig. \ref{fig:classresults_trainratio_hypso-jln-mr} follow a similar pattern. Below a value of around 100 000 samples \mr\ dominates in performance, between 100 000 to 1 000 000 samples \mr\ slightly dominates but for single classes \justoliunet\ achieves better results and above that \justoliunet\ achieves better performance. 
This effect is more apparent in \hyko\ and \hsidrive\ as they are less balanced, resulting in more classes and shares falling below the mentioned coarse sample number. \hypso\ in contrast is mostly balanced and only has three classes, meaning that even when using only a 5\% share of the training data, in almost every case sufficient class samples are available.
Also it can be seen that in general better results are achieved with more samples, which is indicated by the diagonal structure that can be observed in Fig. \ref{fig:classresults_trainratio_hyko-jln-mr} and \ref{fig:classresults_trainratio_hsidrive-jln-mr}. Fig. \ref{fig:classresults_trainratio_hypso-jln-mr} does not show this diagonal structure as there are mostly sufficient samples available to train a robust classifier and performance scores are overall higher than for the other two datasets.

\subsection{HDC-MiniROCKET vs. MiniROCKET} \label{sec:mrhdcmr}
\begin{figure}
    \centering
    \includegraphics[width=\columnwidth]{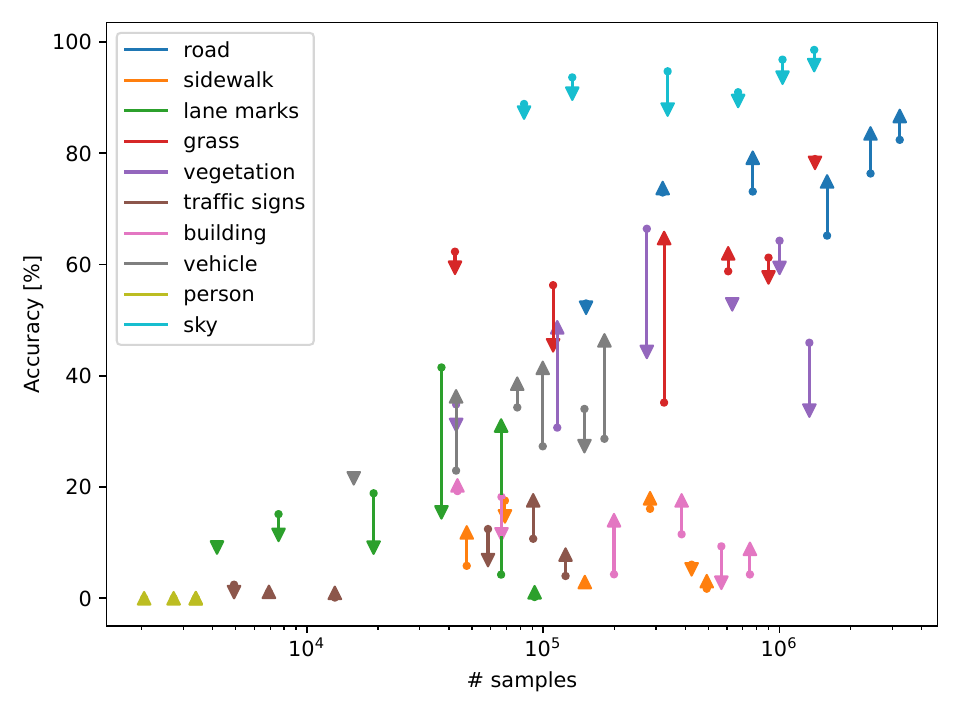}
    \vspace{\negspace}
    \caption{Difference in class accuracy between \mr\ and \hdcmr\ per number of absolute samples in train set shares for \hyko. Arrows point from \mr s accuracy score towards the score for \hdcmr.}
    \label{fig:classresults_trainratio_hyko-mr-hdcmr}
\end{figure}

\begin{figure}
    \centering
    \includegraphics[width=\columnwidth]{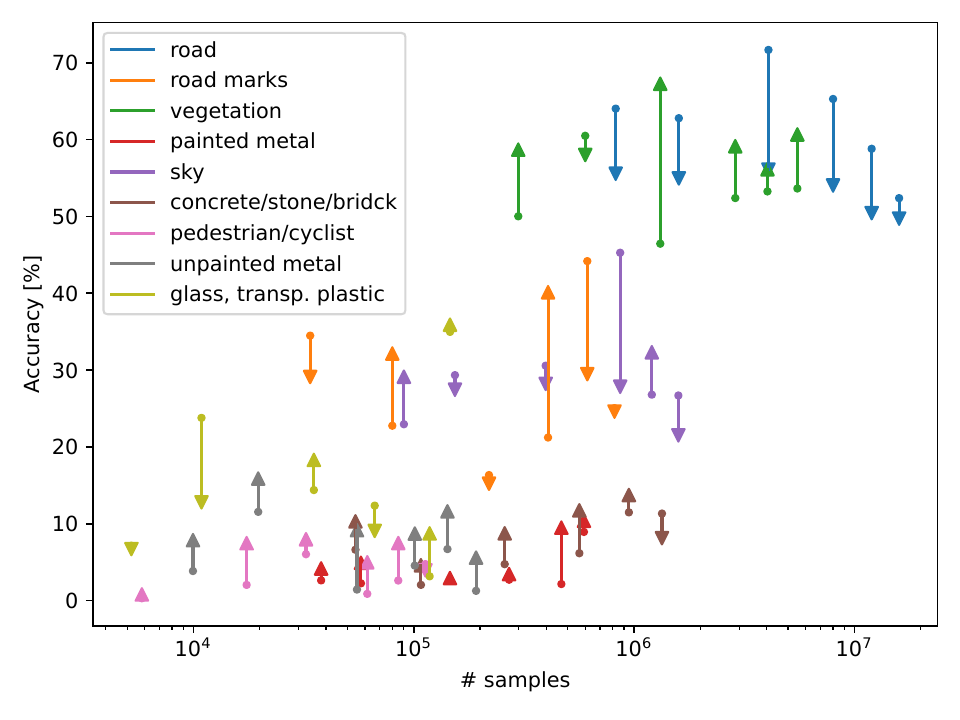}
    \vspace{\negspace}
    \caption{Difference in class accuracy between \mr\ and \hdcmr\ per number of absolute samples in train set shares for \hsidrive. Arrows point from \mr s accuracy score towards the score for \hdcmr.}
    \label{fig:classresults_trainratio_hsidrive-mr-hdcmr}
\end{figure}

\begin{figure}
    \centering
    \includegraphics[width=\columnwidth]{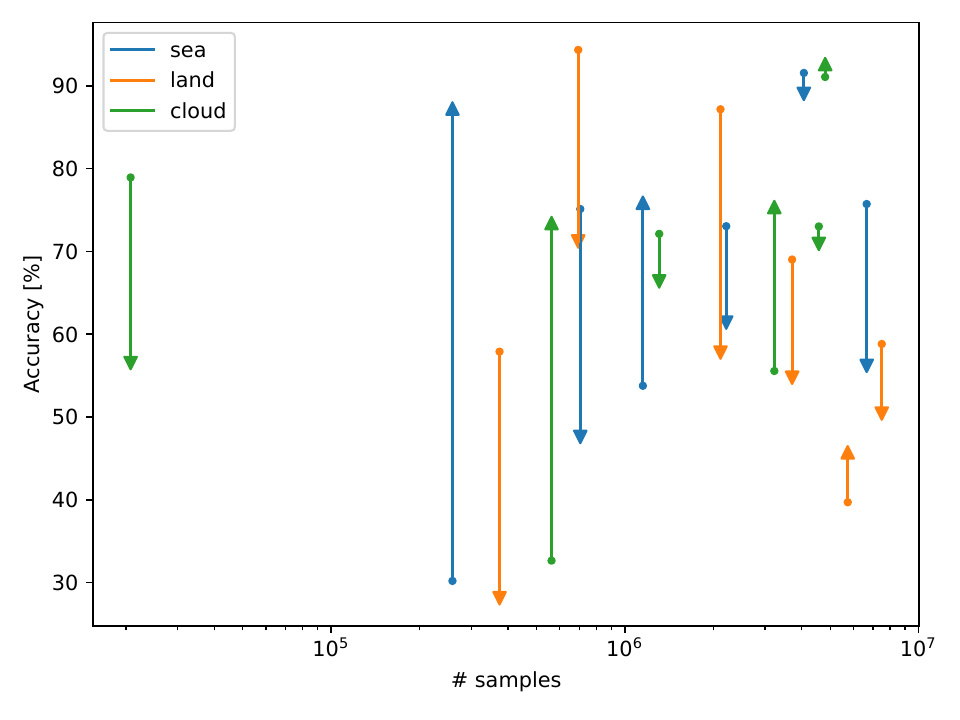}
    \vspace{\negspace}
    \caption{Difference in class accuracy between \mr\ and \hdcmr\ per number of absolute samples in train set shares for \hypso. Arrows point from \mr s accuracy score towards the score for \hdcmr.}
    \label{fig:classresults_trainratio_hypso-mr-hdcmr}
\end{figure}

\hdcmr\ maintains the positional relationship of the elements in the input sequence and encodes them into the feature vector by associating a positional encoding with the intermediate features using \gls{hdc} binding. We assess the impact of this additional information, which is not fully captured with \mr, on model performance. The results are shown in Fig \ref{fig:classresults_trainratio_hyko-mr-hdcmr} to \ref{fig:classresults_trainratio_hypso-mr-hdcmr}. In the plots the arrows point from \mr s performance scores to those of \hdcmr, \ie\ upward arrow meaning \hdcmr\ is better and downward arrow meaning \mr\ is better. 

The \hdcmr\ introduces an additional hyperparameter, scale $s$, into the model. It defines how fast the similarity of the positional encodings change over the length of the input sequence and in turn defines how strongly neighbouring elements of a certain radius are related. If $s=1$ the similarity of the encodings decreases but only reaches zero at the end of the sequence, with $s=2$ the similarity reaches zero after half the sequence and so on \cite{Schlegel2022HET}.

To identify a suitable scale $s$, we performed a preliminary experiment. We train \hyko, \hsidrive, and \hypso\ on 50\% of training data for 10 epochs with different scales $s = \{1, 2, 5, 7, 10, 20, 50, 100\}$ and see which $s$ achieved the maximum validation scores for all metrics. Given the results, for the rest of the experiments $s=5$ was fixed.
%We use only 50\% of training data to reduce the training time. The best $s$ per dataset and metric are shown in Table \ref{tab:scaling}. 
The variation in number of bands and spectral resolution suggests that the best $s$ would vary for each dataset. This is not the case, which might indicate that below a certain bandwidth threshold the signals correlation is so strong that finer bands do not provide additional information for the classification task at hand.


The plots for \hyko\ and \hsidrive\ in Fig. \ref{fig:classresults_trainratio_hyko-mr-hdcmr} and Fig. \ref{fig:classresults_trainratio_hsidrive-mr-hdcmr} again show a diagonal structure, similar to the previous experiments, indicating that more training samples improve model performance.
However, in these experiments a clear improvement for \hdcmr\ over \mr\ can not be observed. While performance for some classes slightly improves, other decline. Overall the data do not show a clear trend. This could have different reasons. First, the convolutions already capture the positional information sufficiently, such that the additional information provided by \hdcmr\ is not relevant for discrimination for the task at hand. This is supported by the fact, that the average spectra for all classes and neighbouring bands are highly correlated. In \cite{Schlegel2022HET} the failure case that is solved by \gls{hdc} are signals that include very sharp peaks at different locations, which is not very common in the spectral signals. Second, it could also be the case that the classifier is not able to capture the additional information and third, that the scale parameter is not optimally selected. A final conclusion needs further analysis and experimentation to better understand \mbox{(HDC-)\mr\ } for spectral classification. 

\subsection{Overall Comparison of Model Performance}

\begin{figure*}
\centering
    \includegraphics[width=0.32\textwidth]{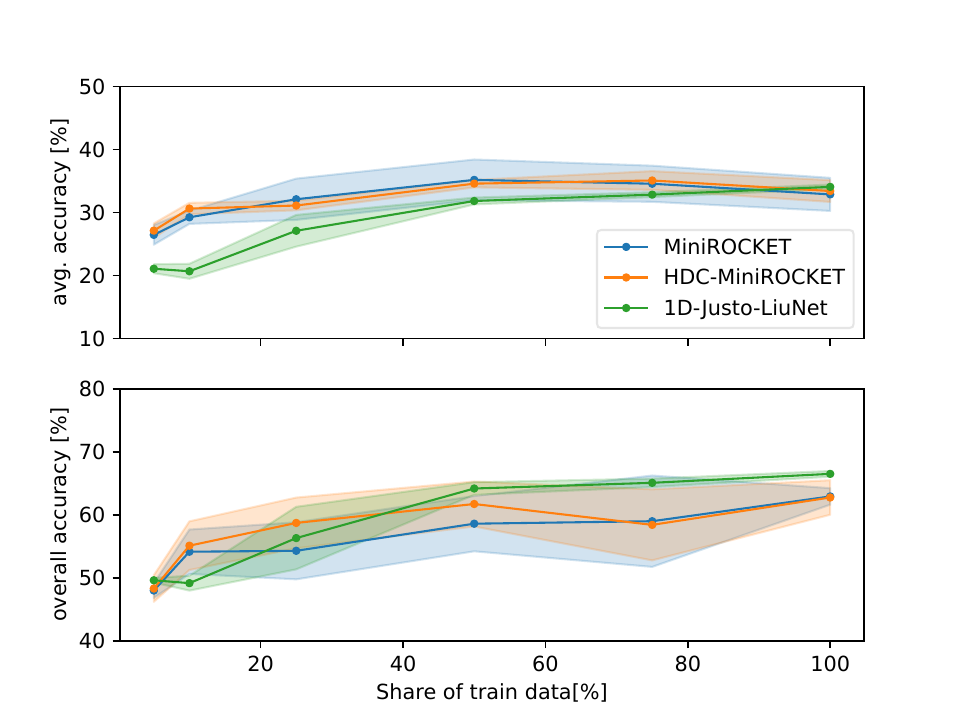}
    \includegraphics[width=0.32\textwidth]{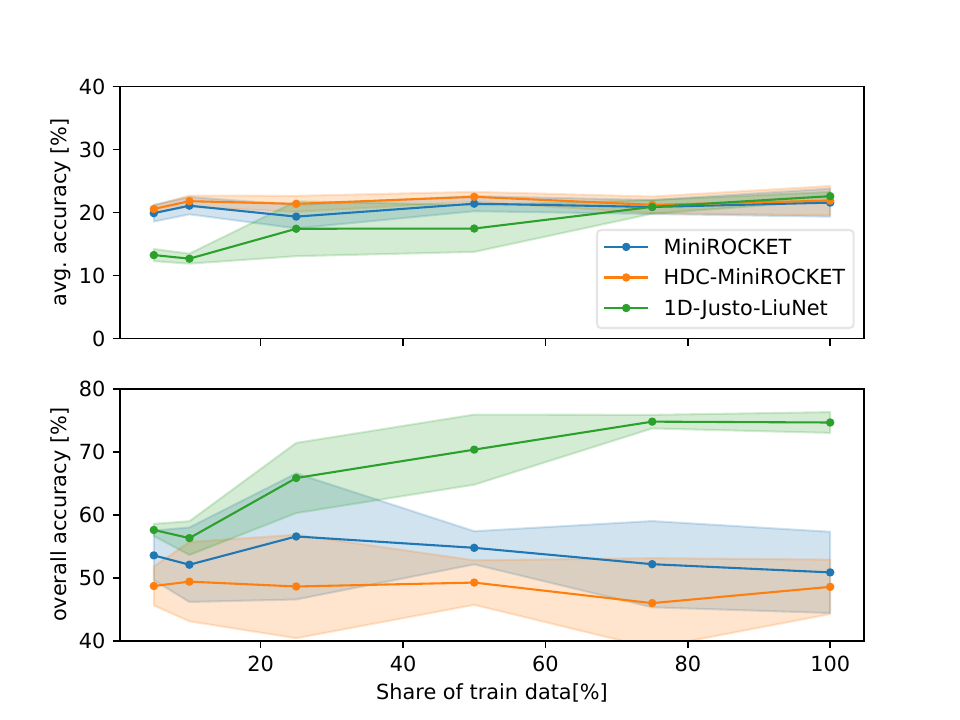}
    \includegraphics[width=0.32\textwidth]{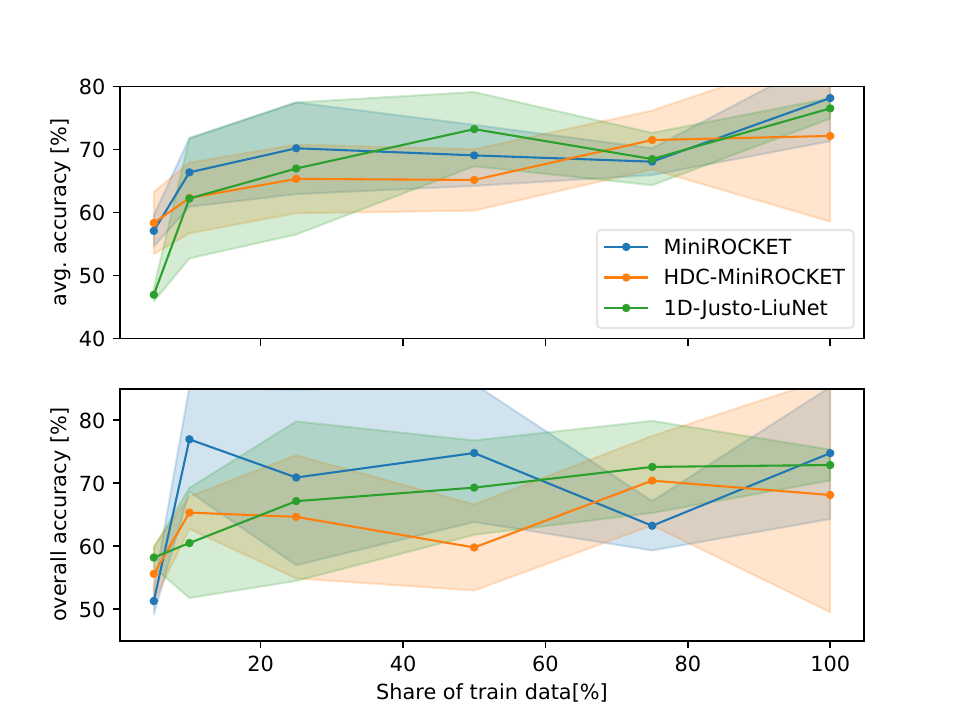}
    \caption{Average and overall accuracy on \hyko\ (left), \hsidrive\ (center) and \hypso\ (right) after training on a share of all training data. Average accuracy is the mean over all class accuracies, whereas overall accuracy averages over all samples.}
    \label{fig:results_aggregated}
\end{figure*}

Finally, we perform a comparison on the overall performance of all models. In Fig. \ref{fig:results_aggregated} we show the overall and average accuracy given the $p\%$ share of the training data. Average accuracy is the mean over all class accuracies, whereas overall accuracy averages over all samples. Hence, with imbalanced datasets the latter is biased towards the performance on majority classes. An overview of the average performance per model per dataset, trained on the full training dataset, is shown in Table \ref{tab:hs3results}.  

What could be observed in plots in sec. \ref{sec:jlnmr} and \ref{sec:mrhdcmr} is confirmed in Fig. \ref{fig:results_aggregated}. The average accuracy shows that \mr\ and \hdcmr\ dominate below a certain value but above that the results are comparable to \justoliunet. For overall accuracy, it can be seen that \justoliunet\ shows better performance, which indicates that \justoliunet\ is better in capturing the distribution of the majority classes while \mr\ is better at capturing all classes. For \hsidrive\ a steady decline in overall performance can be observed. The average accuracy remains stable, which indicates that the \mr\ classifier performance for majority classes suffers from the improvement in performance for minority classes.
Table \ref{tab:hs3results} shows that for the full dataset \justoliunet\ outperforms \mr\ in terms of sample-averaged performance, but has only a slight advantage in class-averaged metrics. The trainable feature extractor gives \justoliunet\ an advantage here, with the trade-off that the model tends to be biased towards majority classes.

The aggregated score are important to assess model performance. However, it should also be noted that it is not sufficient to just look at those summary statistics, as it is often done in the literature. The results can be especially misleading when the distributions are highly imbalanced. Using class- and sample-averaged statistics mitigates this problem, but it does not replace analysing the class-wise performances, as they can reveal patterns that would otherwise remain hidden, \eg\ in Fig. \ref{fig:classresults_trainratio_hyko-jln-mr} it can be seen that the performance for class \textit{sky} is not affected by limited training data as opposed to the class \textit{grass}. 

Finally, we measured the inference time, which showed that for the available implementations \justoliunet\ is a factor of around 40 times faster than \mr. It should, however, be noted that \justoliunet\ uses standard pytorch-modules, while \mr\ is more complex and could probably be further accelerated by code optimization. Additionally, the number of features for \justoliunet\ is much lower than \mr\ with 9996 features. Hence, dimensionality reduction or a smaller set of convolutions could further reduce inference times.

\begin{table}
    \centering
    \footnotesize
    \caption{Results on test data per dataset and model trained on the full training dataset}.%}The 
    \hspace*{-0.7cm}
    \begin{tabular}{lr|cccc}
                    \toprule
                       && \multicolumn{4}{c}{Testing}  \\\cmidrule(lr){3-6}
                       Dataset & Approach & \acrshort{oa} & \acrshort{aa} & $\fone$ & \acrshort{miou} \\
                                \midrule
                       \hyko & \justoliunet & 66.50 & 34.05 & 32.79 & 25.78 \\
                       \hyko & \mr & 62.91 & 32.86 & 30.60 & 24.16\\
                       \hyko & \hdcmr & 62.74 & 33.39 & 31.19 & 25.08 \\
                       \midrule
                       \hsidrive & \justoliunet & 74.65 & 22.58 & 22.31 & 17.67\\
                       \hsidrive & \mr & 50.86 & 21.55 & 18.89 & 14.39\\
                       \hsidrive & \hdcmr & 48.56 & 21.87 & 18.75 & 14.39\\
                        \midrule
                        \hypso & \justoliunet & 72.90 & 76.51 & 69.69 & 56.37 \\
                        \hypso & \mr & 74.79 & 78.15 & 72.27 & 58.91\\
                        \hypso & \hdcmr & 57.75 & 64.97 & 55.91 & 40.98\\
                       \bottomrule
    \end{tabular}
    \label{tab:hs3results}
\end{table}

\section{Conclusion}

In this work we investigate \mr\ and \hdcmr\ for spectral classification and compare it to the state-of-the-art method \justoliunet . Our experiments indicate that \justoliunet\ is preferable for classes with many available training samples and \mr\ is preferable when train data is limited - \ie\ less than around 100 000 samples per class. Furthermore, we showed that \mr s features are general enough to give decent performance over all classes - majority and minority, \justoliunet\ maximized performance mostly on majority classes where sufficient samples are available to robustly train its feature extractor. 
Furthermore, we compared \mr\ and \hdcmr\ to assess if the latter's ability, to capture positional relationships of the input elements, improves classification performance. In our experiments we did not observe a clear improvement. This might indicate that the intrinsic ability of convolutions to capture positional information is already sufficient for the task at hand. It could, however, also be a result of suboptimal selection of the scale hyperparameter. A final answer to this, requires further investigation. 
Additionally, in the future we plan to examine the models performance on larger and more complex datasets, \eg\ \gls{hcv}. In this context, an investigation on dimensionality reduction or reduction of the size of \mr 's fixed kernel set is also very interesting. The bias computation in MiniROCKET, based on a subset of the training data, as it is done now, may be suboptimal when combined with a neural network classifier. As it relies solely on the first batch for bias fitting, the fixed biases may not generalize well to subsequent data. Enhancing bias fitting could further improve the performance of both MiniROCKET and HDC-MiniROCKET.
Finally, we want to highlight the importance of examining the individual class performances -- especially with highly imbalanced datasets, such as the ones used in HS3-Bench. While class- and sample-averaged metrics are useful, it can not completely substitute the analysis of model performance per each class. Hence, HS3-Bench would benefit from including individual class performance.
In conclusion, we recommend to use \justoliunet\ if sufficient training data is available. In the limited data case \mr\ is an effective method for spectral classification that does suffers less from bias towards majority classes than \justoliunet\ does.

\bibliographystyle{IEEEtran}
\bibliography{main}

\end{document}